# Representation of linguistic form and function in recurrent neural networks

Ákos Kádár
a.kadar@uvt.nl

Grzegorz Chrupała
g.chrupala@uvt.nl

Afra Alishahi
a.alishahi@uvt.nl

Tilburg Center for Cognition and Communication
Tilburg University

## Abstract

We present novel methods for analyzing the activation patterns of RNNs from a linguistic point of view and explore the types of linguistic structure they learn. As a case study, we use a multi-task gated recurrent network architecture consisting of two parallel pathways with shared word embeddings trained on predicting the representations of the visual scene corresponding to an input sentence, and predicting the next word in the same sentence. Based on our proposed method to estimate the amount of contribution of individual tokens in the input to the final prediction of the networks we show that the image prediction pathway: a) is sensitive to the information structure of the sentence b) pays selective attention to lexical categories and grammatical functions that carry semantic information c) learns to treat the same input token differently depending on its grammatical functions in the sentence. In contrast the language model is comparatively more sensitive to words with a syntactic function. Furthermore, we propose methods to explore the function of individual hidden units in RNNs and show that the two pathways of the architecture in our case study contain specialized units tuned to patterns informative for the task, some of which can carry activations to later time steps to encode long-term dependencies.

## 1 Introduction

Recurrent neural networks (RNNs) were introduced by Elman (1990) as a connectionist architecture with the ability to model the temporal dimension. They have proved popular for modeling language data as they learn representations of words and larger linguistic units directly from the input data, without feature engineering. Variations of the RNN architectures have been applied in several NLP domains such as parsing (Vinyals et al., 2015) and machine translation (Bahdanau et al., 2015), as well as in computer vision applications such as image generation (Gregor et al., 2015) and object segmentation (Visin et al., 2015). RNNs are also important components of systems integrating Vision and Language, e.g. image (Karpathy and Fei-Fei, 2015) and video captioning (Yu et al., 2015).

These networks can represent variable-length linguistic expressions by encoding them into a fixed-size low-dimensional vector. The nature and the role of the components of these representations are not directly interpretable as they are a complex, non-linear function of the input. There have recently been numerous efforts to visualize deep models such as convolutional neural networks in the domain of computer vision, but much less so for variants of RNNs and for language processing.

The present paper develops novel methods for uncovering abstract linguistic knowledge encoded by the distributed representations of RNNs, with a specific focus on analyzing the hidden activation patterns rather than word embeddings and on the syntactic generalizations that models learns to capture. In the current work we apply our methods to a specific architecture trained on specific tasks, but also provide pointers about how to generalize the proposed analysis to other settings.

As our case study we picked the IMAGINET model introduced by Chrupała et al. (2015). It is a multi-task, multi-modal architecture consisting of two Gated-Recurrent Unit (GRU) (Cho et al., 2014;

Chung et al., 2014) pathways and a shared word embedding matrix. One of the GRUs (VISUAL) is trained to predict image vectors given image descriptions, while the other pathway (TEXTUAL) is a language model. This particular architecture allows a comparative analysis of the hidden activations patterns between networks trained on two very different tasks, while keeping the training data and the word embeddings fixed. Recurrent neural language models akin to TEXTUAL which are trained to predict the next symbol in a sequence are relatively well understood, and there have been some attempts to analyze their internal states (e.g. Elman, 1991; Karpathy et al., 2015). In constrast, VISUAL maps a complete sequence of words to a representation of a corresponding visual scene and is a less commonly encountered, but a more interesting model from the point of view of representing meaning conveyed via linguistic structure.

We report a thorough *macro level* quantitative analysis to provide a linguistic interpretation of the networks' activation patterns. These experiments rely on a novel method we call *omission score* to measure the importance of input tokens to the final prediction of models that compute distributed representations of sentences. This includes various RNN architectures, Recursive Neural Networks and Convolutional Neural Networks. Based on these omission scores our experiments show that the image prediction pathway in general pays special attention to words of syntactic categories that carry semantic content. Furthermore, we observe that on top of lexical features it learns important aspects of the information structure of sentences, and treats the same input tokens differently depending on their grammatical function in the sentence. In contrast, the language model is more sensitive to the local syntactic characteristics of the input sentences.

We also perform a *micro level* analysis to explore the function of individual hidden units by introducing a method we dub *top-k-contexts*. It involves identifying the sentential contexts which yield the highest activation values and the analysis we present is applicable to uni-directional RNN architectures. Through a qualitative examination of these contexts we observe that in case of IMAGINET some of the contexts represent a particular syntactic category or dependency function, while others correspond to a semantic theme. We identify cases where the hidden units encode characteristics of the context that go beyond their lexical properties, and represent abstract patterns that are useful for the network's task. Furthermore, we explore and visualize units that carry their activation over to later time steps to extract longer dependencies and more complex linguistic features. We also quantitatively show that features encoded by the language model are more associated with syntactic constructions than in case of the image prediction pathway.

## 2 Related work

The direct predecessors of modern architectures were first proposed in the seminal paper of Elman (1990). He modifies the recurrent neural network architecture of Jordan (1986) by changing the output-to-memory feedback connections to hidden-to-memory recurrence, enabling Elman networks to represent arbitrary dynamic systems. In Elman (1991) he trains an RNN on a small synthetic sentence dataset and analyzes the activation patterns of the hidden layer. His analysis shows that these distributed representations encode lexical categories, grammatical relationships and hierarchical constituent structures. Giles et al. (1991) trains RNNs similar to Elman networks on strings generated by small deterministic regular grammars with the objective to recognize positive and reject negative strings, and develops the *dynamic state partitioning* technique to extract the learned grammar from the networks in the form of deterministic finite state automatons.

More closely related is the recent work of Li et al. (2015), who develop techniques for a deeper understanding of the activation patterns of RNNs, but focus on models with modern architectures trained on large scale data sets. More specifically, they train Long Short-Term Memory networks (LSTM) (Hochreiter and Schmidhuber, 1997) for phrase level sentiment analysis and present novel methods to explore the inner workings of RNNs. They measure the salience of tokens in sentences by taking the first-order derivatives of the loss with respect to the word embeddings and provide evidence that LSTMs can learn to attend to important tokens in sentences. Furthermore, they plot the activation values of hid-den units through time using heat maps and visualize local semantic compositionality in RNNs. In comparison, the present work focuses more on exploring structure learning in RNNs and on developing methods for a comparative analysis between RNNs.

Adding an explicit attention mechanism that allows the RNNs to focus on different parts of the input was recently introduced by Bahdanau et al. (2015) in the context of extending the sequence-to-sequence RNN architecture for neural machine translation. At the decoding side this neural module assigns weights $\alpha_1, \ldots, \alpha_i$ to the hidden states of the decoder $h_1, \ldots, h_i$, which allows the decoder to selectively pay varying degrees of attention to different phrases in the source sentence at different decoding time-steps. They also provide qualitative analysis by visualizing the attention weigths and exploring the importance of the source encodings at various decoding steps. Similarly Rocktäschel et al. (2016) use an attentive neural network architecture to perform natural language inference and visualize which parts of the hypotheses and premises the model pays attention to when deciding on the entailment relationship. Conversely, the present work focuses on RNNs without an explicit attention mechanism.

Karpathy et al. (2015) also take up the challenge of rendering RNN activation patterns understandable, but use character level language models and rather than taking a linguistic point of view, focus on error analysis and training dynamics of LSTMs and GRUs. They show that certain dimensions in the RNN hidden activation vectors have specific and interpretable functions. One of the goals of the present paper is also to explore the specific functions encoded by individual hidden units, but in a linguistically-informed way.

Li et al. (2016) train a Convolutional Neural Networks (CNN) with different random initializations on the ImageNet dataset (Krizhevsky et al., 2012). For each layer in all networks they store the activation values produced on the validation set of ILSVRC and align similar neurons of different networks. They conclude that while some features are learned across networks, some seem to depend on the initialization. Other works on visualizing the role of individual hidden units in deep models for vision synthesize images by optimizing random images through backpropagation to maximize the activity of units (Erhan et al., 2009; Simonyan et al., 2014; Yosinski et al., 2015; Nguyen et al., 2016) or to approximate the activation vectors of particular layers (Mahendran and Vedaldi, 2015b; Dosovitskiy and Brox, 2015).

In general, there has been a growing interest within computer vision in understanding deep models, with a number of papers dedicated to visualizing learned CNN filters and pixel saliencies (Simonyan et al., 2014; Yosinski et al., 2015; Mahendran and Vedaldi, 2015a). These techniques have also led to improvements in model performance (Eigen et al., 2014) and transferability of features (Zhou et al., 2015). To date there has been much less work on such issues within computational linguistics. We aim to fill this gap by adapting existing methods as well as developing novel techniques to explore the linguistic structure learned by recurrent networks.

## 3 Models

One of the main difficulties for training traditional Elman networks arises from the fact that they overwrite their hidden states at every time step with a new value computed from the current input $x_t$ and the previous hidden state $\mathbf{h_{t-1}}$. Similarly to LSTMs, Gated Recurrent Unit networks introduce a mechanism which facilitates the retention of information over multiple time steps. Specifically, the GRU computes the hidden state at current time step, $\mathbf{h}_t$, as the linear combination of previous activation $\mathbf{h_{t-1}}$, and a new *candidate* activation $\tilde{\mathbf{h}}_t$:

$$\text{GRU}(\mathbf{h}_{t-1}, \mathbf{x}_t) = (1 - \mathbf{z}_t) \odot \mathbf{h}_{t-1} + \mathbf{z}_t \odot \tilde{\mathbf{h}}_t \quad (1)$$

where $\odot$ is elementwise multiplication, and the update gate activation $\mathbf{z_t}$ determines the amount of new information mixed in the current state:

$$\mathbf{z}_t = \sigma_s(\mathbf{W}_z \mathbf{x}_t + \mathbf{U}_z \mathbf{h}_{t-1}) \quad (2)$$

The candidate activation is computed as:

$$\tilde{\mathbf{h}}_t = \sigma(\mathbf{W}\mathbf{x}_t + \mathbf{U}(\mathbf{r}_t \odot \mathbf{h}_{t-1})) \quad (3)$$

The reset gate $\mathbf{r_t}$ determines how much of the current input $\mathbf{x_t}$ is mixed in the previous state $\mathbf{h}_{t-1}$ to form the candidate activation:

$$\mathbf{r}_t = \sigma_s(\mathbf{W}_r \mathbf{x}_t + \mathbf{U}_r \mathbf{h}_{t-1}) \quad (4)$$

## 3.1 Imaginet

IMAGINET introduced in Chrupała et al. (2015) is a multi-modal GRU network architecture that learns visually grounded meaning representations from textual and visual input. It consists of two GRU pathways, TEXTUAL and VISUAL, with a shared word-embedding matrix. The inputs to the model are pairs of image descriptions and their corresponding images. Each sentence is mapped to two sequences of hidden states, one by VISUAL and the other by TEXTUAL:

$$\mathbf{h}_t^V = \text{GRU}^V(\mathbf{h}_{t-1}^V, \mathbf{x}_t) \quad (5)$$
$$\mathbf{h}_t^T = \text{GRU}^T(\mathbf{h}_{t-1}^T, \mathbf{x}_t) \quad (6)$$

At each time step TEXTUAL predicts the next word in the sentence $S$ from its current hidden state $\mathbf{h}_t^T$, while VISUAL predicts the image-vector[1] $\hat{\mathbf{i}}$ from its last hidden representation $\mathbf{h}_t^V$.

$$\hat{\mathbf{i}} = \mathbf{V}\mathbf{h}_\tau^V \quad (7)$$
$$p(S_{t+1}|S_{1:t}) = \text{softmax}(\mathbf{L}\mathbf{h}_t^T) \quad (8)$$

The loss function is a multi-task objective which penalizes error on the visual and the textual targets simultaneously. The objective combines cross-entropy loss $L^T$ for the word predictions and cosine distance $L^V$ for the image predictions.[2]

$$L^T(\theta) = -\frac{1}{\tau}\sum_{t=1}^{\tau}\log p(S_t|S_{1:t}) \quad (9)$$

$$L^V(\theta) = 1 - \frac{\hat{\mathbf{i}} \cdot \mathbf{i}}{\|\hat{\mathbf{i}}\|\|\mathbf{i}\|} \quad (10)$$

$$L = \alpha L^T + (1-\alpha)L^V \quad (11)$$

For more details about the model and its performance see Chrupała et al. (2015).

---

[1] Representing the full image, extracted from the pre-trained Convolutional Neural Network of (Simonyan and Zisserman, 2015).

[2] Our version of the model has two minor modifications compared the the original IMAGINET: we use cosine distance as the visual loss, and we use image vectors where each dimension is transformed by subtracting the mean and dividing by standard deviation.

## 4 Experiments

In the following sections, we report a series of experiments in which we explore the kinds of linguistic regularities the TEXTUAL and VISUAL pathways of IMAGINET learn from word-level input. Section 5 presents the macro-level analyses where we propose methods to analyze the final hidden activation vectors of the recurrent pathways from linguistic point of view. Section 6 reports exploratory experiments on the linguistic features encoded by individual hidden units. In both sections we report our findings based on the IMAGINET model and discuss the generalizabilty of our methods to other architectures. For all the experiments, we trained IMAGINET on the training portion of the MSCOCO image-caption dataset (Lin et al., 2014), and analyzed the representations of the sentences in the validation set. The target image representations were extracted from the pre-softmax layer of the 16-layer CNN (Simonyan and Zisserman, 2015).

## 5 Analysis of hidden activation vectors

In this section we propose novel techniques for interpreting the activation patterns of neural networks trained on language tasks from a linguistic point of view and focus on the high level understanding of what types of words the networks pay most attention to. Furthermore, we investigate if the networks learn to assign appropriate amounts of importance to tokens depending on their position and grammatical function in the sentences. In Section 5.1 we introduce *omission scores*; a metric to measure the contribution of each token to the prediction of the networks. Section 5.2 aggregates the omission scores in terms of dependency relations and part-of-speech categories and compares VISUAL and TEXTUAL. Lastly Section 5.3 investigates the extent to which the importance of words for the different pathways depend on the words themselves, their position and their grammatical function in the sentences.

### 5.1 Computing Omission Scores

In both pathways of IMAGINET the full sentences are represented by the activation vector at the end-of-sentence symbol ($\mathbf{h}_\text{end}$). We measure the salience of each word $S_i$ in an input sentence $S_{1:n}$ based on how much the representation of the partial sentence $S_{\setminus i} = S_{1:i-1}S_{i+1:n}$, with the omitted word $S_i$, deviates from that of the original sentence representation. For example, the distance between $\mathbf{h}_{\text{end}}(\textit{the black dog is running})$ and $\mathbf{h}_{\text{end}}(\textit{the dog is running})$ determines the importance of *black* in the first sentence. We introduce the measure $\text{omission}(i, S)$ for estimating the salience of a word $S_i$:

$$\text{omission}(i, S) = 1 - \text{cosine}(\mathbf{h}_{\text{end}}(S), \mathbf{h}_{\text{end}}(S_{\setminus i})) \quad (12)$$

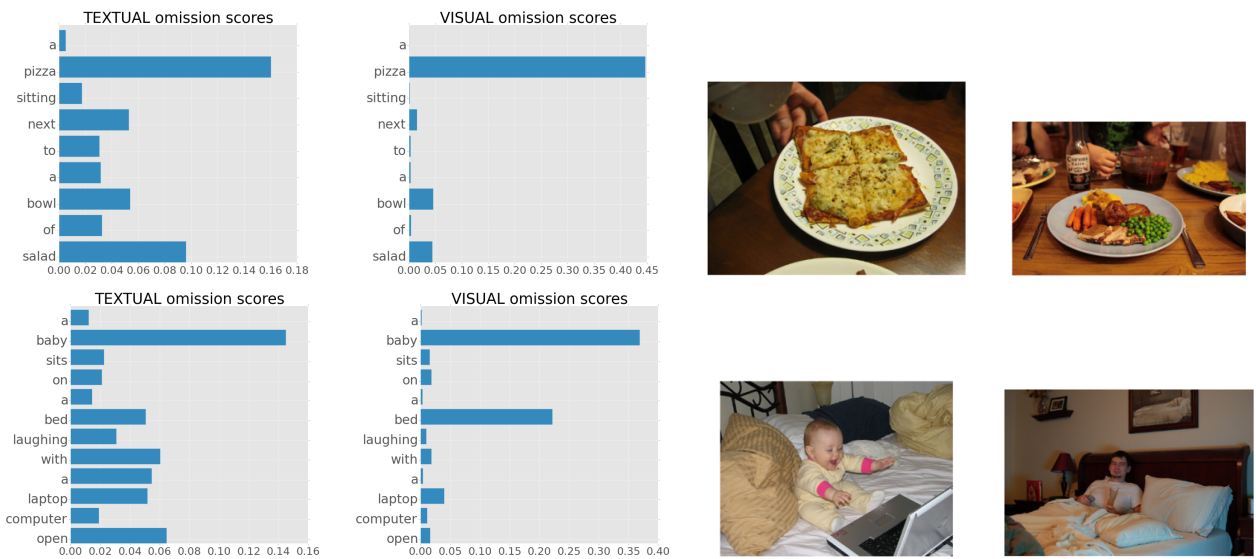

Figure 1: The omission scores of VISUAL and TEXTUAL for two example sentences, and the best retrieved images for the original sentence (left) and the sentence with the most important word removed (right).

Figure 1 demonstrates the omission measure for the VISUAL and TEXTUAL pathways for two example captions. For both captions the omission scores are plotted along with the first image retrieved by VISUAL for the full sentence and for the sentence with the word with the highest $\text{omission}(i, S)$ removed. The images are retrieved from the validation set of MS-COCO by 1) computing the image representation of the given sentence with VISUAL 2) extracting the CNN features for the images from the set 3) finding the image that minimizes the cosine distance to the query. In the first example, the omission scores for VISUAL suggest that the model interpreted *pizza* as the most important word in the sequence and returned an image that depicts a pizza on a plate. Removing the word *pizza* promotes *salad* and *bowl* as the main theme of the sentence and the model retrieves an image with a dining table with salad-like dishes. For the second example, the omission scores for VISUAL show that the model paid attention mostly to *baby* and *bed* and slightly to *laptop* and retrieved an image depicting a baby sitting on a bed with a laptop. Removing the word *baby* leads to an image that depicts an adult male laying on a bed. Figure 1 also shows that in contrast to VISUAL, in both examples TEXTUAL distributes its attention more evenly across time steps instead of focusing on the types of words related to the corresponding visual scene.

For other RNN architectures such as GRUs, LSTMs and their bi-directional variants to measure the contribution of tokens to their predictions the omission can be straight-forwardly computed using their hidden state at the last time step used for prediction. Furthermore, the technique can be applied in general to other architectures, which embed variable length linguistic expressions to the same fixed dimensional space and perform predictions based on these embeddings. This includes tree-structured Recursive Neural Network models such as the Tree-LSTM introduced in (Tai et al., 2015) or the CNN architecture of (Kim, 2014) for sentence classification. In both cases the pre-softmax activations can be extracted from the models as the representations of the full and partial sentences.

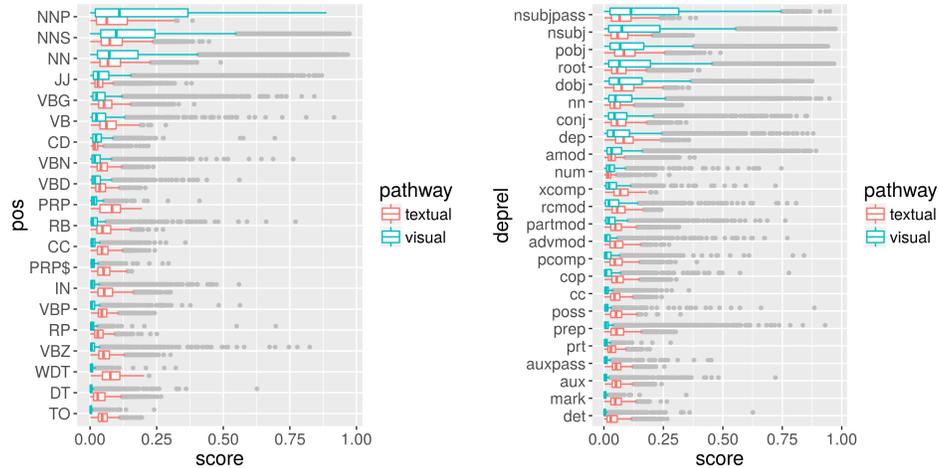

Figure 2: Distribution of omission scores for POS (left) and deprel labels (right), for the TEXTUAL and VISUAL pathways. Only labels which occur at least 500 times are included.

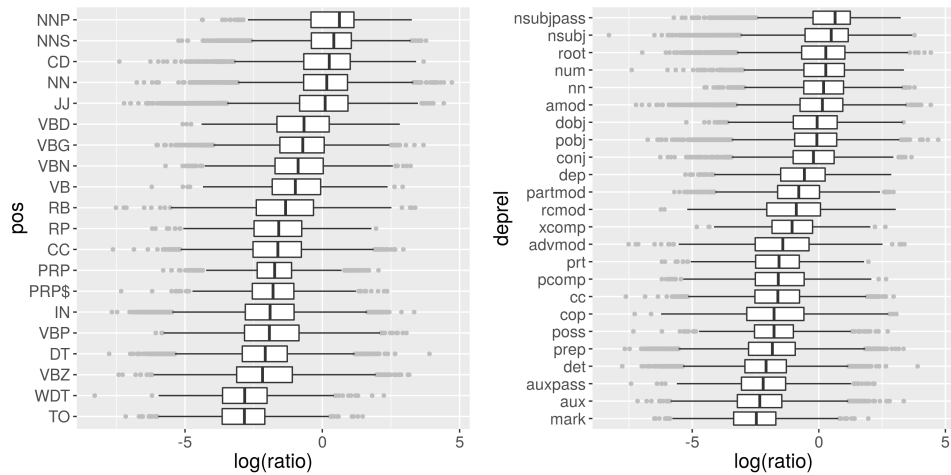

Figure 3: Distributions of log ratios of omission scores of TEXTUAL to VISUAL per POS (left) and deprel labels (right). Only labels which occur at least 500 times are included.

## 5.2 Omission score distributions

The OMISSION scores can be used not only to estimate the importance of individual words, but also of syntactic categories. We estimate the salience of each syntactic category by accumulating the omission scores for all words in that category. We tag every word in a sentence with the part-of-speech (POS) category and the dependency relation (deprel) label of its incoming arc. For example, for the sentence *the black dog*, we get (*the*, DT, det), (*black*, JJ, amod), (*dog*, NN, root). Both POS tagging and dependency parsing are performed jointly using the TurboParser dependency parser (Martins et al., 2013).[3] The POS tags used are the Penn Treebank tags and the dependencies are the Stanford basic dependencies.

Figure 2 shows the distribution of omission scores per POS and deprel label for the two pathways of IMAGINET. The general trend is that for the VISUAL pathway, the omission scores are high for a small subset of labels - corresponding mostly to nouns, less so for adjectives and even less for verbs - and low for the rest (mostly function words and various types of verbs). For TEXTUAL the differences are smaller, and the pathway seems to be sensitive to the omission of most types of words. Figure 3 compares the two pathways directly using the log of the ratio of the VISUAL to TEXTUAL omission scores, and plots the distribution of this ratio for different POS and deprel labels. Log ratios above zero indicate stronger association with the VISUAL pathway and below zero with the TEXTUAL pathway. We see that in relative terms, VISUAL is more sensitive to nouns (NNP, NNS, NN), numerals (CD) and adjectives (JJ), and TEXTUAL to the prepositions (TO, IN), some types of verbs (VBZ, VBP), determiners (DET, WDT) and particles (RP). This picture is complemented by the analysis of the relative importance of dependency relations: VISUAL pays most attention to the relations NSUBJPASS, NSUBJ, POBJ, ROOT, DOBJ, NN, CONJ, DEP, AMOD and NUM, whereas TEXTUAL is more sensitive to DET, MARK, AUX, AUXPASS, PRT, PREP, POSS, CC and COP. As expected, VISUAL is more focused on grammatical functions typically filled by semantically contentful words, while TEXTUAL distributes its attention more uniformly and attends relatively more to purely grammatical functions. It is worth noting, however, the relatively low omission scores for verbs in case of VISUAL. One might expect that the task of image prediction from descriptions requires general language understanding and so high omission scores for all content words in general, however, the results suggest that this setting is not optimal for learning useful representations of verbs, which possibly leads to representations that are too task specific and not transferable accross tasks.

---
[3] Available at github.com/andre-martins/TurboParser.

### 5.3 Beyond Lexical Cues

Models that utilize the sequential structure of natural-language have the capacity to interpret the same word-type differently depending on the context. The omission score distributions in Section 5.2 show that in the case of IMAGINET the pathways are differentially sensitive to content vs. function words. This may be either just due to purely lexical features or the model may actually learn to pay more attention to the same word type in appropriate contexts. This section investigates to what extent VISUAL and TEXTUAL discriminate between occurrences of a given word in different positions and grammatical functions. The analysis we described here takes the omission scores as input data, therefore it can be potentially applied to any architecture for which the omission scores can be computed. However, the presented analysis and results regarding word positions can only be meaningful for Recurrent Neural Networks as they compute their representations sequentially and are not limited by fixed windowsizes.[4]

We fit four linear models which predict the omission scores per token with the following predictor variables:

1. LM WORD: word type
2. LM DEPREL: word type, dependency label and their interaction
3. LM POSITION: word type, position (binned as first, second, third, middle, antepenult, penult, last) and their interaction
4. LM FULL: word type, dependency label, position, word-deprel interaction, word-position interaction

We split the whole MSCOCO validation set into two parts, fit the models (regularized via L2 penalty) on the first part, and compute the proportion of variance explained on the second part. For comparison we also train a version of IMAGINET, where the GRU in the VISUAL pathway is replaced by a simple vector-summation operation over word-embeddings and thus it does not have access to sequential cues. This model we refer to as SUM. Figure 4 shows the increase in $R^2$ for the models per pathway relative to LM WORD. The plot reveals that adding additional information on top of lexical features in case of SUM does increase the explained variance slightly, which is probably due to the unseen words in the held out set. More interestingly the relative increase in $R^2$ between LM WORD and LM FULL in case of VISUAL is larger, suggesting that the omission scores in case of VISUAL do not only depend on the words themselves, but their grammatical function and position in sentences. The raw $R^2$ scores show that for the TEXTUAL model a) the word-type predicts the omission-score to much smaller degree compared to VISUAL (0.38 against 0.48), and the increase in $R^2$ shows that the position of the word adds consider-able amount of information.

---

[4]CNNs with multi-word filters and tree-structured recursive neural networks do not incrementally build representations of sentences in a left-to-right or right-to-left fashion. Bi-directional RNNs, however, are affected by word-order and can potentially learn to handle the same word in different positions differently.

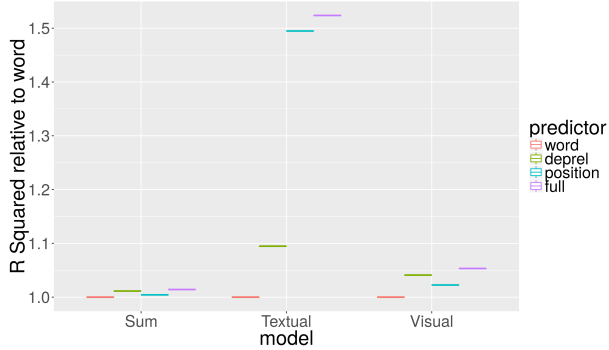

Figure 4: Proportion of variance in omission scores explained by the models for SUM, VISUAL and TEXTUAL pathways.

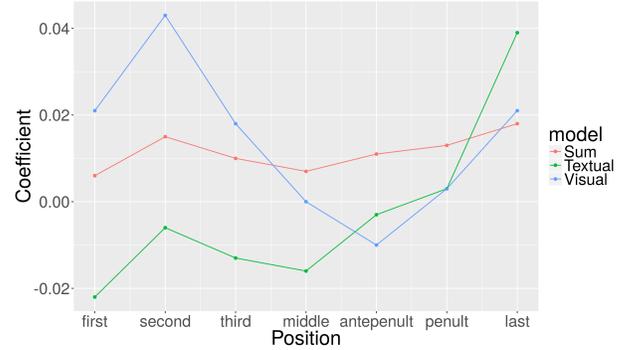

Figure 6: Coefficients on the y-axis of LM FULL corresponding to the position variables on the x-axis.

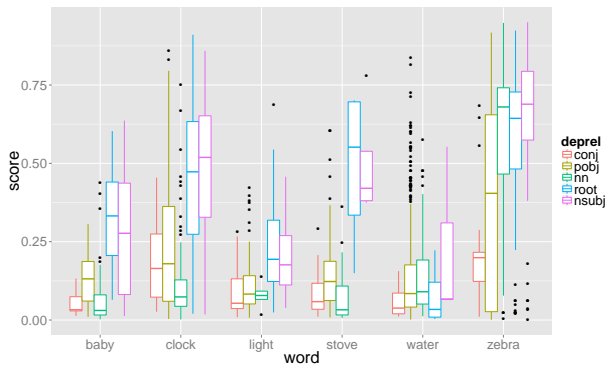

Figure 5: Distribution of omission scores per deprel label for the selected word types.

### 5.3.1 Sensitivity to grammatical function

In order to find out some of the specific configurations leading to in increase in $R^2$ in case of VISUAL, we next considered all word types with occurrence counts of at least 100 and ranked them according to how much better, on average, LM DEPREL predicted their omission scores compared to LM WORD. Figure 5 shows the per-dependency omission-score distributions for the five top ranked words plus the word *water*. There are clear and large differences in how these words impact the network's representation depending on what grammatical function they fulfill. They all have large omission scores when they occur as NSUBJ (nominal subject) or ROOT, likely due to the fact that these grammatical functions typically have a large contribution to the complete meaning of a sentence. Conversely, all have small omission scores when appearing as CONJ (conjunct): this is probably because in this position they share their contribution with the first, often more important, member of the conjunction (e.g. *A cow and its baby eating grass*). The pattern for NN (nominal modifier) is a bit more complicated: for four of the words shown (as well as for most other words not shown in the figure), the score is very low in this grammatical function–presumably because most words contribute less to the sentence meaning when used as modifiers than as heads (e.g. *a clock tower*). However, for the words *zebra* and *water*, omission scores are high when they act as a nominal modifier NN. This appears due to two reasons:

1. For *zebra*: there are frequent erroneous parses such as *zebra/*NN *browsing/*ROOT instead of *zebra/*NSUBJ *browsing/*ROOT. The network does not make this mistake, and treats these occurrences of *zebra* according to its importance as NSUBJ.

2. *Water* as a modifier often changes the meaning of its head in a visually salient way: e.g. *water fall, water balloon, water scene, water skiing*, and thus the network learns that this particular word is important in the modifier position.

### 5.3.2 Sensitivity to information structure

As observed in Section 5.3 adding extra information about the position of words explains more of the variance than in case of VISUAL and especially TEXTUAL. Figure 6 shows the coefficients corresponding to the position variables in LM FULL. Since the omission scores are measured at the end-of-sentence token, the expectation is that for TEX-

TUAL, as a language model, the words appearing closer to the end of the sentence would have a stronger effect on the omission scores. This seems to be confirmed by the plot as the coefficients until the *penultimate* are all negative. For the VISUAL model it is less clear what to expect: on the one hand due to their chain structure, RNNs are better at keeping track of short-distance rather than long-distance dependencies and thus we can expect tokens in positions closer to the end of the sentence to be more important. On the other hand in English the information structure (or pragmatic structure) of a sentence is expressed via linear ordering: the TOPIC of a sentence appears sentence-initially, and the COMMENT follows. In the context of other text types such as dialog or multi-sentence narrative structure, we would expect COMMENT to often be more important than TOPIC as COMMENT will often contain new information in these cases. In our setting of image captions however, it is the TOPIC that typically contains the most important objects depicted in the image, e.g. *two zebras are grazing in tall grass on a savannah*. Thus, for the task of predicting features of the visual scene, it would be advantageous to detect the topic of the sentence and up-weight its importance in the final meaning representation. Figure 6 appears to support this hypothesis and the network does learn to pay more attention to words appearing sentence-initially. This effect seems to be to some extent mixed with the recency bias of RNNs as perhaps indicated by the relatively high coefficient of the *last* position for VISUAL.

## 6 Analysis of hidden units

### 6.1 Top $K$ contexts

The aim of this section is to develop methods that allow for the qualitative analysis of the kinds of linguistic features individual hidden dimensions of RNNs encode. We develop a simple method we dub *top K contexts* after the *top K images* of Zhou et al. (2015). This involves forwarding each sentence from a corpus token-by-token through an RNN, and storing the hidden activation of the network $\mathbf{h}_t$ for each time step $t$. This results in an activation matrix $M \in R^{d \times n}$, where $d$ is the number of hidden dimensions and $n$ is the total number of time steps (or tokens) in the whole corpus. Each cell $M_{it}$ in the resulting matrix represents the activation value of the $i^{\text{th}}$ unit for some token at time step $t$ in the corpus. Making the assumption that high activation values indicate importance, we sort the rows of the activation matrix $M$ by the magnitude of the activations, leading to the top $K$ contexts for each unit. This method can be straight-forwardly applied to various RNN architectures such as Elman networks or LSTMs as it only requires storing the activation values for hidden units and their corresponding context. For architectures with $n$ hidden layers one could extract multiple activation matrices $M^1, \ldots, M^n$ and perform analysis on each of them separately. In the following sections 6.2, 6.3 and 6.4 we provide a qualitative exploration of the kinds of features the VISUAL and TEXTUAL pathways of IMAGINET encode based on analysis on the *top K contexts* from the validation portion of the MS-COCO data set. All techniques introduced in these section can be applied in the same setting where *top K contexts* can be applied. A limitation of the generalizability of our analysis is that in the case of bi-directional architectures the interpretation of the features extracted by the RNNs that process the input tokens in the reversed order might be hard from a linguistic point of view.

Table 1: Contexts from the top 20 trigrams for example hidden units in each pathway.

| VISUAL | TEXTUAL |
|---|---|
| and a laptop, cables on it, camera parts and, and cables on, cords and cables | other cars driving, engine car traveling, watches cars racing, with passengers driving |
| crowd together on, team run on, group of men, of men behind, team crowd together | a sandy beach, lush green hillside, of a beach, a rocky hillside, a dry river |
| with broccoli carrots, with noodles corn, bowl with meat, salad has broccoli, salad with broccoli | tomatoes on a, food on a, broccoli on a, vegetables on a, vegetables on a |
| two teddy bears, teddy bears posing, three teddy bears, teddy bears sitting, bears sitting next | tennis court waiting, water and waiting, table and ready, cooked and ready, into a waiting |
| a dirt track, a race track, a stunt jump, stunt jump in, in the air | on the shore, on the floor, at the end, at the shore, on the side |

### 6.2 Specialized hidden units

Table 1 shows the top 5 trigram contexts with the highest activations for five example hidden units for VISUAL and TEXTUAL. It shows the general pat-tern that the individual dimensions become highly sensitive towards contexts with syntactically and/or topically related patterns. For example the top 20 trigram contexts for the first hidden unit of VISUAL in Table 1 all contain tokens topically related to home electronics, such as phones, remotes and camera parts. Top-20 5-gram contexts for this unit include: *cell phone calculator and gum, all hanging on wires like, such as beads and cords*. More interestingly, the first hidden unit for TEXTUAL in Table 1 seems to be highly active for a combined syntactic and semantic template: contexts including a token corresponding to a vehicle followed by a transportation verb. Exploring a larger context of 5-grams reveals other interesting units with high activations for such semantic/syntactic constructions in TEXTUAL, e.g. *a dog pokes his head, white cat sticking his head, a dog sticking its head, a dog sticking his head, with a long tail perched*.

### 6.3 Units predictive of a grammatical function

To explore the syntactic functions encoded by specialized dimensions, we train two logistic regression models (one for VISUAL and one for TEXTUAL) to predict the dependency label of a token at time step $t$. The models use two sets of predictors:
- hidden activation vectors $\mathbf{h}_t^V$ or $\mathbf{h}_t^T$
- n-gram features up to a window size of 4; for example, to predict the label for *dog* in the sentence *the nice dog* we extract $the_2$, $nice_1$, $dog_0$, $the_2\ nice_1$, $nice_1\ dog_0$, $the_2\ nice_1\ dog_0$, etc.

For both VISUAL and TEXTUAL, we pinpoint the hidden units that are predictive of grammatical function by taking the top 5 logistic regression coefficients $\beta^V$ and $\beta^T$ per dependency label corresponding to the dimensions of $\mathbf{h}_t^V$ and $\mathbf{h}_t^T$ with the highest absolute value. Since the models include n-gram predictors, the logistic regression model will only have coefficients with high absolute values for units that are predictive of dependency relation over and above the n-gram features, and thus which most likely represent some type of functional information.

Table 2 shows the top four context representations for the hidden units corresponding to one of the top 5 highest coefficients for a number of the deprels for both VISUAL and TEXTUAL. Some of the units in Table 2 seem to offer solely lexical cues to the deprel prediction model, while others encode more general syntactic information. A number of units have high activations for target words typically fulfilling the same grammatical function:

- The example unit for the category POSS in case of TEXTUAL has top contexts with target words *his, her* and *their*. This is also true for VISUAL.
- The example unit for NUM for TEXTUAL has high activations for both *two* and *three*.
- The example units for CC given for VISUAL has high activations in the presence of target tokens *and* and *or*.
- The contexts for the dimension of VISUAL with the highest coefficient for AUX have both *is* and *are* as target tokens.

Table 2: Examples from the top 20 contexts for one of the top five most predictive dimensions per deprel label in TEXTUAL and VISUAL, using the combinations of n-gram features and activation vectors.

| TEXTUAL | VISUAL |
|---|---|
| **poss**: dog sticking his, is brushing her, child brushing their, is brushing his | **aux**: tennis player is, racquet that is, colored ties are, but they are |
| **num**: several other hot, woman holding two, playing with two, group of three | **cc**: a waterway and, for construction or, some rocks and, with trees and |
| **advmod**: up a very, in a very, on a very, others watch very | **num**: pole in two, light with two, steeple and two, set between two |
| **pobj**: a table covered, at a baseball, in a baseball, at a professional | **prep**: area surrounded by, a field with, and mountains in, the background among |
| **advcl**: serious while playing, standing and playing, couch while playing, as he plays | **poss**: court holding her, player holds his, to come his, ball with her |

### 6.4 Units carrying over information

Further examination of some of the highly activated units reveals dimensions that are predictive of deprels that require information about the identities or grammatical functions of previous tokens. For example, predictive units for POBJ in the TEXTUAL pathway in Table 2 generalize over the prepositions *at* and *in*, and contexts in the top 20 include additional prepositions such as *around a dirt* and *on a grass*. But interestingly, rather than being active for the prepositions themselves, all top 20 contexts belong to the construction PREP DET POBJ where the object has to do with outdoors. For VISUAL, one of the top units for the category CONJ has top contexts *with greenery and, the table and, colored*

*circles and, wooden furniture and, furniture and a*. Given that the value of this dimension predicts the presence of a conjunct at the current time step, this particular dimension seems to carry over its high activation value to the next time step, since all the 5 example trigrams require a conjunct in the next step.

This is also the case for the most predictive units of VISUAL for POBJ: the top contexts for this unit are *several bunches of, with lots of, the end of, has trays of* and *in front of*, suggesting that the information content of the token *of* must be carried over the next time step.

To visually explore the phenomenon of units carrying over information through time steps, we searched for interesting hidden units in VISUAL using their top-20 5-gram representations, and plotted their activation values through time for some example captions. We only used example sentences where the activation of the hidden unit was in the highest decile. Figure 7 shows the results. The first two rows are examples of *lexicalized units* recognizing topically related words and keeping them in memory until the end of the sequence. The next two rows demonstrate a hidden unit active for the multi-word expressions *next to a* and *next to an*. The following three rows show a unit active for noun phrase constructions which contain a numeral followed by a reference to a person. The last three examples show a dimension that has a modest activation for tokens of category FOOD, but has a high activation for a following food item accompanying it in the visual scene. In the very last example, the unit has a modest activation for the token *broccoli*, then its activation decreases for *on a*. With the arrival of the token *plate* the activation increases again, has even higher activation for *with* and finally the highest for *potatoes*. The top 20 5-grams for this hidden unit all contain multiple food items, such as vegetables and meat with chopsticks.

## 6.5 Comparison of models based on top contexts

The results in Section 5 highlight some of the differences between TEXTUAL and VISUAL. We saw that VISUAL learns to pay relatively more attention to contentful words and TEXTUAL to words with purely grammatical function. Moreover, while for TEXTUAL tokens appearing near the end of the sen-tence are more salient in general, VISUAL learns to pay attention to sentence-initial nouns as they are likely the TOPIC of the sentence. In this section we explore a further comparison between the models based on the hypothesis that due to their different objectives, the activity of the hidden dimensions of VISUAL are more characterized by semantic relationships between contexts, whereas the dimensions of TEXTUAL are more focused on extracting syntactic patterns. In order to test this hypothesis quantitatively, we measure the strength of association between activations of hidden units and either lexical (token n-grams) or structural (deprel n-grams) types of contexts.

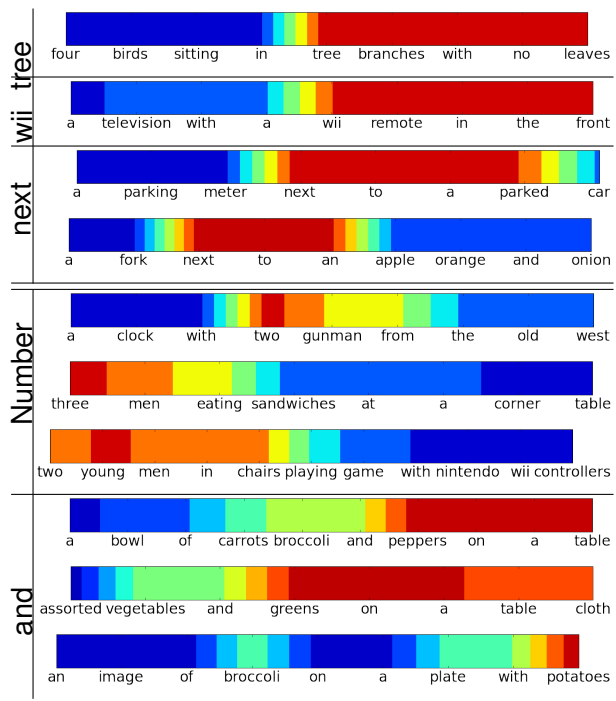

Figure 7: Hidden units of VISUAL active for meaningful constructions. The red end of the spectrum corresponds to higher activation values of the hidden unit.

We define $A_j^m$ as a discrete random variable corresponding to a binned activation over time steps of model $m$ at hidden dimension $j$, and $C$ as a discrete random variable indicating the context (where type of context can be for example word trigram or deprel bigram). The strength of association between $A_j^m$ and $C$ can be measured by their mutual infor-

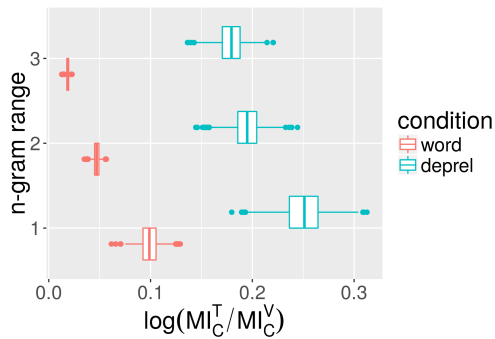

Figure 8: Bootstrap distributions of log ratios of median mutual information scores for word and deprel contexts.

mation:

$$\mathrm{I}(A_j^m; C) = \sum_{a \in A_j^m} \sum_{c \in C} p(a,c) \log\left(\frac{p(a,c)}{p(a)p(c)}\right)$$

Similarly to Li et al. (2016), the activation value distributions are discretized into percentile bins per dimension, such that each bin contains 5% of the marginal density. For context types, we used dependency label and word uni-, bi- and trigrams. For simplicity we use the notation $\mathrm{MI}_C^m$ to denote the median mutual information score over all units of the pathway $m$ when considering context $C$.

We then compute log ratios $\log(\mathrm{MI}_C^T/\mathrm{MI}_C^V)$ for all six context types $C$. In order to quantify variability we bootstrap this statistic with 5000 replicates. Figure 8 shows the resulting bootstrap distributions for uni-, bi-, and trigram contexts, in the word and deprel conditions. The clear pattern is that the log ratios are much higher in case of deprels, with no overlaps between the bootstrap distitributions. Thus, in general, the size of the relative difference between TEXTUAL and VISUAL median mutual information score is much more pronounced for deprel context types. This suggests that features that are encoded by the hidden units of the models are indeed different, and that the features encoded by TEXTUAL are more associated with syntactic constructions than in case of VISUAL.

## 7 Conclusion

The goal of our paper is to propose novel methods for the analysis of the encoding of linguistic knowledge in RNNs trained on language tasks. We focused on developing quantitative methods to measure the importance of different kinds of words to the decision of such models. Furthermore, we proposed techniques to explore what kinds of linguistic features the models learn to exploit on top of lexical cues. Using the IMAGINET model as our case study our analyses of the hidden activation patterns show that the VISUAL model learns an abstract representation of the information structure of the language, and pays selective attention to lexical categories and grammatical functions that carry semantic information. In contrast, the language model TEXTUAL is sensitive to features of a more syntactic nature. We have also shown that each network contains specialized units which are tuned to both lexical and structural patterns that are useful for the task at hand, some of which can carry activations to later time steps to encode long-term dependencies. In future we would like to apply the techniques introduced in this paper to analyze the encoding of linguistic form and function of recurrent neural models trained on different objectives, such as neural machine translation systems (e.g. Sutskever et al., 2014) or the purely distributional sentence embedding system of Kiros et al. (2015). A number of recurrent neural models rely on a so called attention mechanism, first introduced by Bahdanau et al. (2015) under the name of soft alignment. In these networks attention is explicitly represented and it would be interesting to see how our method of discovering implicit attention, the omission score, compares. For future work we also propose to collect data where humans assess the importance of each word in sentences and explore the relationsship between *omission scores* for various models and human annotations. Finally, one of the benefits of understanding how linguistic form and function is represented in RNNs is that it can provide insight into how to improve systems. We plan to draw on lessons learned from our analyses in order to develop models with better general-purpose sentence representations.